\pdfoutput=1

\documentclass[11pt]{article}

\usepackage{acl}

\usepackage{times}
\usepackage{latexsym}
\usepackage{times}  
\usepackage{helvet}  
\usepackage{courier}  
\usepackage{graphicx} 
\usepackage{natbib}  
\usepackage{caption} 

\usepackage{algorithm}
\usepackage{algorithmic}
\usepackage{booktabs}  
\usepackage{multirow}

\usepackage{xcolor}
\definecolor{Purple}{RGB}{192, 172, 217}
\definecolor{Gray}{RGB}{153, 153, 153}
\definecolor{Blue}{RGB}{0, 113, 188}
\definecolor{SeaGreen}{RGB}{67, 205, 128}
\definecolor{GoldNrod}{RGB}{218, 165, 32}
\usepackage{pifont} 
\newcommand{\okmark}{{\textbf{\textcolor[rgb]{0.1, 0.5, 0.1}{$\checkmark$}}}}
\newcommand{\ngmark}{{\textbf{\color{red}{\ding{55}}}}}

\usepackage{amssymb}

\usepackage[T1]{fontenc}

\usepackage[utf8]{inputenc}

\usepackage{microtype}

\title{Fact-driven Logical Reasoning for Machine Reading Comprehension}

\author{Siru Ouyang\textsuperscript{1}, Zhuosheng Zhang\textsuperscript{2}, Hai Zhao\textsuperscript{2} \\
  \textsuperscript{1} University of Illinois Urbana-Champaign, \textsuperscript{2} Shanghai Jiao Tong University   \\
  \normalsize\texttt{siruo2@illinois.edu, zhangzs@sjtu.edu.cn, zhaohai@cs.sjtu.edu.cn}}

\begin{document}
\maketitle
\begin{abstract}
Recent years have witnessed an increasing interest in training machines with reasoning ability, which deeply relies on accurately and clearly presented clue forms. The clues are usually modeled as entity-aware knowledge in existing studies. However, those entity-aware clues are primarily focused on commonsense, making them insufficient for tasks that require knowledge of temporary facts or events, particularly in logical reasoning for reading comprehension. To address this challenge, we are motivated to cover both commonsense and temporary knowledge clues hierarchically. Specifically, we propose a general formalism of knowledge units by extracting backbone constituents of the sentence, such as the subject-verb-object formed ``facts''. We then construct a supergraph on top of the fact units, allowing for the benefit of sentence-level (relations among fact groups) and entity-level interactions (concepts or actions inside a fact). Experimental results on logical reasoning benchmarks and dialogue modeling datasets show that our approach improves the baselines substantially, and it is general across backbone models. Code is available at \url{https://github.com/ozyyshr/FocalReasoner}.
\end{abstract}

\section{Introduction}

Training machines to understand human languages is a long-standing goal of artificial intelligence \citep{hermann2015teaching}, with a wide range of application scenarios such as question-answering and dialogue systems. As a well-established task, machine reading comprehension (MRC) has attracted research interest for a long time. MRC challenges machines to answer questions based on a referenced passage. \citep{chen2016thorough,sachan2016machine,Seo2016Bidirectional,dhingra2017gated,Cui2017Attention,song2018exploring,hu2019read,zhang2019sg,back2020neurquri,zhang2020retrospective}. There has been remarkable progress in MRC, with human-parity benchmark results reported in examination-style MRC datasets like SQuAD \citep{rajpurkar-etal-2016-squad} and RACE \citep{lai2017race}. 

\begin{figure*}[htb]
\centering
\includegraphics[width=0.98\textwidth]{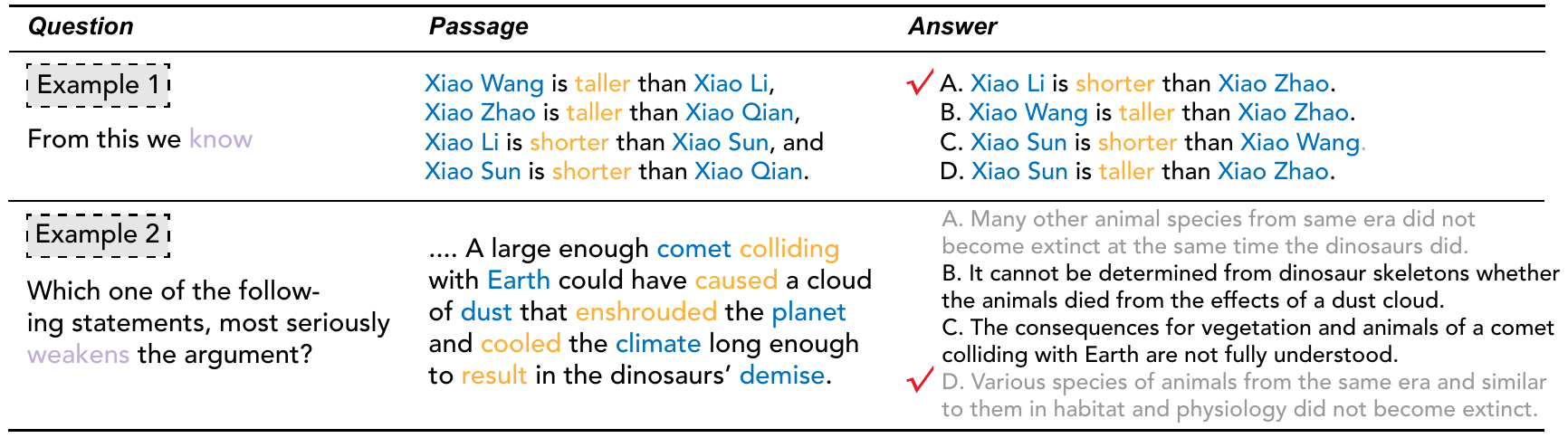}
\caption{Two examples from LogiQA and ReClor respectively are illustrated. There are arguments and relations between arguments. Both are emphasized by different colors: \textcolor{Blue}{arguments}, \textcolor{GoldNrod}{relations}. Keywords in questions are highlighted in \textcolor{Purple}{Purple}. Key options are highlighted in \textcolor{Gray}{gray}. }
\label{example_logic}
\end{figure*}

However, the extent to which systems have grasped the required knowledge and reading comprehension skills remains a concern \citep{sugawara2020assessing}. A recent trend is to decompose comprehension ability into a collection of skills, such as span extraction, numerical reasoning, commonsense reasoning, and logical reasoning. Among these tasks, logical reasoning MRC has shown to be particularly challenging. It requires the machine to examine, analyze and critically evaluate arguments based on the relationships of facts that occur in ordinary languages, instead of simple pattern matching as focused in earlier studies \citep{lai2021machine}. ReClor \cite{yu2020reclor} and LogiQA \cite{ijcai2020-0501} are two representative datasets introduced to promote the development of logical reasoning in MRC, where their questions (Figure \ref{example_logic}) are selected from standardized exams such as GMAT and LSAT\footnote{\url{https://en.wikipedia.org/wiki/Law_School_Admission_Test}}.

Recent studies typically exploit a pre-trained language model (PLM) as a key encoder for effective contextualized representation. However, according to diagnostic tests \citep{ettinger2020bert,rogers2020primer}, PLMs like BERT \citep{devlin-etal-2019-bert}, despite encoding syntactic and semantic information through large-scale pre-training, they tend to struggle with understanding role reversal and pragmatic inference, as well as role-based event knowledge. Therefore, studies show that PLMs perform poorly in logical reasoning MRC tasks \citep{yu2020reclor,ijcai2020-0501}, as the supervision required for these tasks is rarely available during pre-training.

Following the research trend of previous reasoning tasks such as commonsense reasoning, a natural interest is to model the entity-aware relationships (e.g., \textit{isA} or \textit{hasA} predicates) in the passages using graph networks  \citep{yasunaga2021qagnn,ren2020beta,zhang2021video,krishna2020local,lv2020graph}. However, for tasks like logical reasoning of text that involve deductive/inductive/abductive reasoning \cite{reichertz20044}, the text usually contains hypothetical conditions that cannot be represented solely by commonsense knowledge. Therefore, these methods may insufficiently capture necessary logical units for inducing answers, since they paay little attention to non-entity, non-commonsense clues \citep{2021arXiv210406598Z}. Referring to example 2 in Figure \ref{example_logic}, previous methods may only consider commonsense knowledge such as ``\textit{Earth is a planet}'', without considering the temporary fact\footnote{A fact can be seen as an observed event \cite{peterson1997fact}.} like ``\textit{comet caused dust}''.

To mitigate the challenge, we are motivated to bridge the gap between commonsense and temporary knowledge. First, we propose a general formalism of knowledge units by extracting backbone constituents of sentences such as the subject-verb-object formed ``facts''\footnote{The definition follows \citet{nakashole2014language}. For example, those units may reflect the facts of \textit{who did what to whom}, or \textit{who is what}}. We then develop the \textsc{Focal Reasoner}, a \underline{f}act-driven l\underline{o}gi\underline{cal} reasoning model, which builds supergraphs on top of these fact units. This approach not only captures the entity-level relations inside a fact unit within supernodes, but also enhances information flow among fact units at the sentence level through holistic supergraph modeling.

\begin{table*}
\centering
\setlength{\belowcaptionskip}{2 pt}
\small
\setlength{\tabcolsep}{9 pt}
\begin{tabular}{lcccc}
    \toprule
    Model    & Knowledge Format & Modeling Method & Entity-level & Sentence-level\\
    \midrule
    LReasoner \cite{wang2021logic} &entities&manual rules/executor&\okmark& \ngmark \\
    DAGN \cite{zhang2021video} & EDUs & graph &\ngmark &\okmark\\
    HGN \cite{chen-etal-2022-modeling-hierarchical} & EDUs and phrases & graph & \ngmark & \okmark\\
    MERIt \cite{jiao2022merit} & entities & graph &\okmark& \ngmark \\
    \midrule
    Ours & fact units & supergraph &\okmark &\okmark \\
    \bottomrule
  \end{tabular}
  \caption{Comparison between our approach \textsc{Focal Reasoner} and previous methods on different aspects.}\label{related-work}
\vspace{-5mm}
\end{table*}

Our model is evaluated on two challenging logical reasoning benchmarks including ReClor\citep{yu2020reclor}, LogiQA\citep{ijcai2020-0501}, and one dialogue reasoning dataset Mutual, for verifying the effectiveness and the generalizability across different domains and question formats. To sum up, our contributions are three folds: 

(i) We propose a general formalism to support representing logic units using backbone constituents of the sentences, as fine-grained knowledge carriers for logical reasoning. 

(ii) We design a hierarchical fact-driven approach to construct a supergraph on top of our newly defined fact units. It models both the sentence-level (relations among fact groups) and entity-level (concepts or actions inside a fact) interactions.

(iii) Empirical studies verify the general effectiveness of our method on logical reasoning for QA and dialogues, with dramatically superior results over baselines. Analysis shows that our method can uncover complex logical structures with supergraph modeling on fact units.

\section{Related Work}
\label{gen_inst}

\subsection{From Machine Reading Comprehension to Reasoning} 
Recent years have witnessed massive research on Machine Reading Comprehension (MRC) whose goal is training machines to understand human languages, which has become one of the most important areas of NLP \citep{chen2016thorough,sachan2016machine,Seo2016Bidirectional,dhingra2017gated,Cui2017Attention,song2018exploring,hu2019read,zhang2019sg,back2020neurquri,zhang2020retrospective}. Despite the success of MRC models on various datasets such as CNN/Daily Mail \citep{hermann2015teaching}, SQuAD \citep{rajpurkar-etal-2016-squad}, RACE \citep{lai-etal-2017-race} and so on, researchers began to rethink what extent does the problem been solved. Nowadays, there is massive research into the reasoning ability of machines. According to \citep{kaushik-lipton-2018-much, ZHOU2020275, chen-etal-2016-thorough}, reasoning abilities can be broadly categorized into (i) commonsense reasoning \citep{davis2015commonsense,bhagavatula2019abductive,talmor2019commonsenseqa,huang2019cosmos}; (ii) numerical reasoning \citep{dua2019drop}; (iii) multi-hop reasoning \citep{yang2018hotpotqa} and (iv) logical reasoning \citep{yu2020reclor,ijcai2020-0501}, among which logical reasoning is essential in human intelligence but has merely been delved into. Natural Language Inference (NLI) \citep{bowman2015large,williams2018broad,nie2020adversarial} is a task closely related to logical reasoning. However, it has two obvious drawbacks in measuring logical reasoning abilities. One is that it only has three logical types which are \textit{entailment, contradiction} and \textit{neutral}. The other is its limitation on sentence-level reasoning. Hence, it is important for comprehensive and deeper logical reasoning abilities.


\subsection{Logical Reasoning of Text}
Neural and symbolic approaches have been explored in logical reasoning of text \cite{garcez2015neural, garcez2022neural, ren2020beta}. Compared with neural methods, symbolic ones such as \cite{wang2021logic} heavily rely on data-specific patterns that are pre- and manually defined. It also suffers from error propagation and unscalable searching spaces. Our method is more related to the neural research line. 

As shown in Table \ref{related-work}, our work mainly differs from previous work in knowledge format and modeling. \citet{zhang2021video} uses discourse relations and designs the discourse-aware graph network to help logical reasoning. HGN \cite{chen-etal-2022-modeling-hierarchical} further leverages key phrases to build both inter-sentence and intra-sentence interactions in the context. The most recent work MERIt \cite{jiao2022merit} builds meta paths among logical variables (consists of entities and phrases) to model the logical relations. For knowledge formats, our proposed \textit{fact units} can better represent both commonsense knowledge and temporary knowledge existing in the context. Previous methods either use elementary discourse units (EDUs) or phrases for temporary knowledge only or named entities to represent ``isA''/``hasA'' for commonsense knowledge only. For modeling methods, \textsc{Focal Reasoner} is able to jointly model sentence-level and entity-level interactions via supergraphs, whereas previous methods capture either entity-level or sentence-level information with simple graph networks. 


\begin{figure*}
\vspace{-3 mm}
\centering
\includegraphics[width=0.9\textwidth]{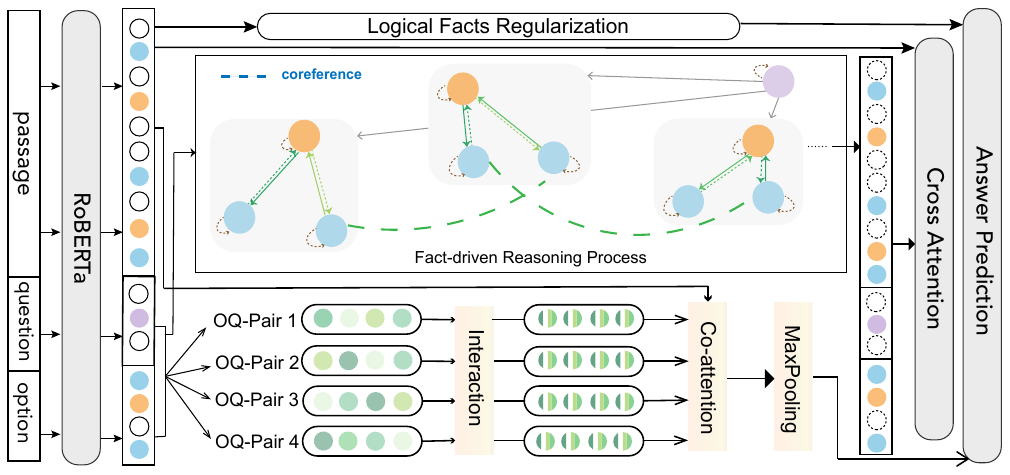}
\caption{The framework or our model. For supergraph reasoning, in each iteration, each node selectively receives the message from the neighboring nodes to update its representation. The dashed circle means zero vector.}
\label{overview}
\vspace{-3 mm}
\end{figure*}

\section{Methodology}
\label{headings}

This section presents our fact-driven approach, \textsc{Focal Reasoner}. The overall architecture is shown in Figure \ref{overview}. \textsc{Focal Reasoner} consists of three stages. Firstly, it extracts fact units from raw texts via syntactic processing and constructs a supergraph. Then, it performs reasoning over the supergraph along with a logical fact regularization. Finally, it aggregates the learned representation to decode the right answer.

\subsection{Fact Unit Extraction and Supergraph Construction}
\label{preprocessing}
\paragraph{Fact Unit Extraction.} The first step is to fetch triplets that constitute a fact unit. To keep the framework generic, we use a fairly simple fact unit extractor based on syntactic relations. Given a context consisting of multiple sentences, we first conduct dependency parsing on each sentence using off-the-shelf tools like SpaCy \citep{spacy2}. After that, we extract the subject, the predicate, and the object tokens to get the \textit{``Argument-Predicate-Argument''} fact units corresponding to each sentence in the context.



\begin{figure*}
\vspace{-3 mm}
\centering
\includegraphics[width=0.9\textwidth]{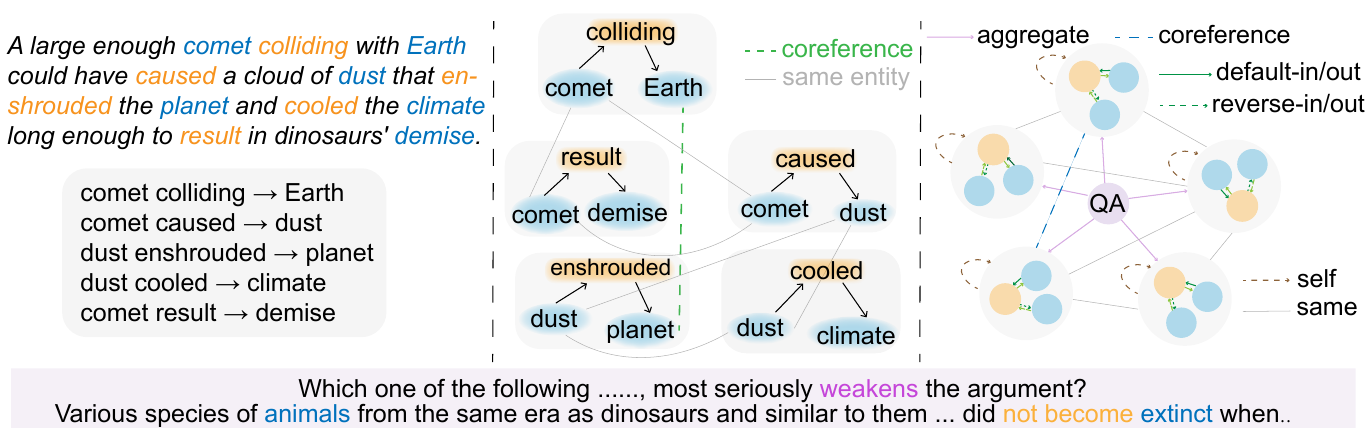}
\caption{The process of constructing the fact chain and its corresponding Levi graph form of an example in Figure \ref{example_logic}. Entities and relations are illustrated in their corresponding color.}
\label{fact_chain}
\vspace{-3 mm}
\end{figure*}

\paragraph{Supergraph Construction.}
With the obtained fact units, 
we construct a super graph as shown in Figure \ref{fact_chain}.
Concretely, the fact units are organized in the form of Levi graph \citep{levi-1942}, which turns arguments and predicates all into nodes. An original fact unit is in the form of $F=(V,E,R)$, where $V$ is the set of the arguments, $E$ is the set of edges connected between arguments, and $R$ is the relations of each edge which are predicates here. The corresponding Levi graph is denoted as $F_l = (V_L, E_L, R_L)$ where $V_L=V \cup R$, which makes the originally directly connected arguments be intermediately connected via relations. $E_L$ is the edges connected between $V_L$. As for $R_L$, previous works such as \citep{marcheggiani-titov-2017-encoding, beck-etal-2018-graph} designed three types of edges $R_L=\{\textup{default}, \textup{reverse}, \textup{self}\}$ to enhance information flow. Here in our settings, we extend it into five types: \textit{default-in, default-out, reverse-in, reverse-out, self}, corresponding to the directions of edges towards the predicates. Detailed descriptions can be found in Appendix \ref{edge_type}.

We construct the supergraph by making connections between fact units $F_l$. In particular, we take three strategies according to global information, identical concept, and co-reference information:

(i) We add a node $V_g$ initialized with the question-option representation and connect it to all the fact unit nodes. The edge type is set as \textit{aggregate} for better information interaction. 

(ii) There can be identical mentions in different sentences, resulting in repeated nodes in fact units. We connect nodes corresponding to the same non-pronoun arguments by edges with edge type \textit{same}. 

(iii) We conduct co-reference resolution on context using an off-to-shelf model\footnote{\url{https://github.com/huggingface/neuralcoref}.} in order to identify arguments in fact units that refer to the same one. We add edges with type \textit{coref} between them. The final supergraph is denoted as $S = (F_l\cup V_g, E_g)$ where $E_g$ is the set of edges added with the previous three strategies.

\subsection{Reasoning Process}
\paragraph{Graph Reasoning.}A natural way to model the supergraph is via Relational Graph Convolution Networks \citep{10.1007/978-3-319-93417-4_38}. For a multiple-choice logical reasoning problem that consists of a context ($C$), a question ($Q$) and an option ($O$), we first concatenate $C$, $Q$, and $O$ to form the input sequence. Then, the input sequence is fed to a pre-trained language model to obtain the encoded representations. We initialize the nodes with averaged hidden states of its tokens because our triplets extraction performs at the word level. For edges, we use a one-hot embedding layer to encode the relations.

Based on the relational graph convolutional network  and given the initial representation $h_i^0$ for every node $v_i$, the feed-forward or the message-passing process with information control can be written as
\begin{equation}
    h_i^{(l+1)}=\textup{ReLU}(\sum_{r\in R_L}\sum_{v_j\in \mathcal{N}_r(v_i)}g_{q}^{(l)}\frac{1}{c_{i,r}}w_r^{(l)}h_j^{(l)}),
\end{equation}
where $\mathcal{N}_r(v_i)$ denotes the neighbors of node $v_i$ under relation $r$ and $c_{i,r}$ is the number of those nodes. $w_r^{(l)}$ is the learnable parameters of layer $l$. $g_q^{(l)}$ is a gated value between 0 and 1. Through the graph encoder $F_G(.)$, we then obtain the hidden representations of nodes in fact units as $\{h_0^F,...h_m^F\} = F_G(\{v_{L,0},...v_{L,m}\},E_L)$ where $h_i^F$ is the node representation inside the fact unit. They are then concatenated as the representation for supernode as $h_0^S$. Therefore, we have $\{h_0,...h_m\} = F_G(\{h_0^S,...h_m^S\},E_g)$

For node features on the supergraph, it is fused via the attention and gating mechanisms with the original representations of the context encoder $H^C$. We apply the attention mechanism to append the supergraph representation to the original one $\tilde{H} = \textup{Attn}(H^c,K_f,V_f)$,
where $\{K_f,V_f\}$ are packed from the learned representations of the supergraph, i.e., $\{h_0,...h_m\}$, and $Attn$ is multi-head self-attention. We compute $\lambda\in[0,1]$ to weigh the expected importance of supergraph representation of each source word $\lambda_1 = \sigma(W_\lambda \tilde{H}+U_\lambda H^C)$,
where $W_\lambda$ and $U_\lambda$ are learnable parameters. $H^C$ and $\tilde{H}$ are then fused for an effective representation $H = H^C+\lambda\tilde{H}\in \mathbb{R}^{4\times d}$.

\begin{table*}[htb]
\vspace{-3mm}
\small
\centering\centering\setlength{\tabcolsep}{9.5pt}
\begin{tabular}{lcccccc}
\toprule
\multirow{2}{*}{Model} &
\multicolumn{4}{c}{ReClor} & \multicolumn{2}{c}{LogiQA}\\
\cmidrule{2-5}
\cmidrule{6-7}
 & Dev & Test & Test-E & Test-H & Dev & Test \\ 
\midrule
Human Performance*& - & 63.0 & 57.1 & 67.2  & -&86.0 \\
\midrule
RoBERTa*  &62.6&55.6&75.5&40.0 & 35.0&35.3  \\
DAGN*  &65.8&58.3&75.9&44.5&36.9&39.3 \\
\quad w/o data augmentation &65.2&58.2&76.1&44.1 & 35.5&38.7\\
LReasoner$\dagger$ & 66.2 & 62.4 &81.4 & 47.5 & 38.1 & 40.6 \\
\quad w/o data augmentation & 65.2 & 58.3 & \textbf{78.6} & 42.3 & - & - \\
MERIt$\dagger$ & 66.8&59.6&78.1&45.2&40.0&40.3 \\
\quad w/o data augmentation & 63.0&57.9&-&-&-&- \\
HGN$\dagger$ &66.4 & 58.7 & 77.7 & 43.8 & 40.1 & 39.9 \\
\textsc{Focal Reasoner}&\textbf{66.8} ($\uparrow$4.2) &\textbf{58.9}($\uparrow$3.3) &77.1($\uparrow$1.6) & \textbf{44.6} ($\uparrow$4.6) &\textbf{41.0} ($\uparrow$6.0) &\textbf{40.3} ($\uparrow$5.0)\\
\midrule
DeBERTa*  & 74.4 & 68.9 & 83.4 & 57.5 & 44.4 & 41.5 \\
LReasoner$\dagger$ &74.6 & 71.8 & 83.4 & 62.7 & 45.8 & 43.3 \\
HGN$\dagger$ &76.0 & 72.3 & 84.5 & 62.7 & 44.9 & 44.2 \\
MERIt$\dagger$ &78.0&73.1&86.2&\textbf{64.4}&-&-\\
\textsc{Focal Reasoner}&\textbf{78.6} ($\uparrow$4.2) & \textbf{73.3} ($\uparrow$4.4)&\textbf{86.4}($\uparrow$3.0)&63.0 ($\uparrow$5.5) & \textbf{47.3} ($\uparrow$2.9)&\textbf{45.8} ($\uparrow$4.3)\\

\bottomrule
\end{tabular}
\caption{Experimental results of our model compared with baseline models on ReClor and LogiQA dataset. Segment-1: Human performance; Segment-2: RoBERTa-based models; Segment-3: DeBERTA-based models. Test-E and Test-H denote Test-Easy and Test-Hard respectively. The results in \textbf{bold} are the best performance except for the human performance. * indicates that the results are taken from \citet{yu2020reclor} and \citet{ijcai2020-0501}. Results with $\dagger$ are taken from their corresponding papers. \textbf{Note} that we are mainly comparing with previous literature without data augmentation (DA), as we hope to concentrate on our research problem on model architecture and logic relation discovery, instead of using additional tricks to bother the attention.}\label{table:e2e}
\end{table*}

\begin{table*}[htb]
\small
\centering\centering\setlength{\tabcolsep}{10.2pt}
\begin{tabular}{lccccccc}
\toprule
\multirow{2}{*}{Model}
& \multicolumn{3}{c}{Dev Set} & \multicolumn{3}{c}{Test Set}  \\

\cmidrule{2-7}
& $R_4@1$ & $R_4@2$ & MRR & $R_4@1$ & $R_4@2$ & MRR \\ 
\midrule
RoBERTa*  &69.5&87.8&82.4&71.3&89.2&83.6\\
RoBERTa-MC* &69.3&88.7&82.5&68.6&88.7&82.2 \\
\midrule
\textsc{Focal Reasoner}&\textbf{73.4} ($\uparrow$4.1) &\textbf{90.3} ($\uparrow$1.6) &\textbf{84.9} ($\uparrow$2.4)&\textbf{72.7} ($\uparrow$4.1) &\textbf{91.0} ($\uparrow$2.3) &\textbf{84.6} ($\uparrow$2.4)\\
\bottomrule
\end{tabular}
\caption{Experimental results of our model compared with baseline on MuTual dataset. * indicates that the results are taken from \cite{mutual}. For a fair comparison with our method, we also report the multi-choice method (RoBERTa-MC) in addition to the default Individual scoring method (RoBERTa).}\label{mutual}
\vspace{-3mm}
\end{table*}

\begin{table*}[htb]
\vspace{-3mm}
\small
\setlength{\belowcaptionskip}{8.3pt}

\centering\centering\setlength{\tabcolsep}{10.3pt}
\begin{tabular}{lcccccccccc}
\toprule
Model  & S & W & I & CMP  & ER & P & D& R & IF &MS \\ 
\midrule
RoBERTa&61.7& 47.8  & 39.1 & 63.9  & 58.3 & 50.8 & 50.0 & 56.3 & 61.5 & 56.7\\
DAGN  & 63.8 & 46.0 & 39.1 & 69.4 & 57.1 & 53.9 & 46.7  & 62.5 & 62.4 & 56.7\\
\midrule
\textsc{Focal Reasoner} &72.3&66.4  & 47.8 & 91.7  & 76.2 & 76.9 & 66.7  & 68.8 & 73.5  & 86.7\\
\bottomrule
\end{tabular}
\caption{Accuracy on the dev set of ReClor on several representative question types. \textit{S: Strengthen, W: Weaken, I: Implication, CMP: Conclusion/Main Point, ER: Explain or Resolve, D: Dispute, R: Role, IF: Identify a Flaw, MS: Match Structures}. All results are reported on the same PLM RoBERTa.}\label{class}
\vspace{-3mm}
\end{table*}

\paragraph{Interaction.} For the application to the concerned QA tasks that require reasoning, options have their inherent logical relations, which can be leveraged to aid answer prediction. Inspired by \citep{ran2019option}, we use an attention-based mechanism to gather option correlation information. 

Specifically for an option $O_i$, the information it gets by interaction with option $O_j$ is calculated as
    $O_i^{(j)} = [O_i^q-\tilde{O}_i^{j};O_i^q\circ \tilde{O}_i^{j}]$,
where $O_i^q$ is the representation of the concatenation for the $i$-th option and question after the context encoder; $\tilde{O}_i^{j} = O_i^qAttn(O_i^q,O_j^q; v)$. Then the option-wise information is gathered to fuse the option correlation information
\begin{equation}
    \hat{O_i}=\tanh (W_c[O_i^q;\{O_i^{(j)}\}_{i\neq j}]+b_c)
\end{equation}
where $\textbf{W}_c\in \mathbb{R}^{d\times 7d}$ and $b_c\in\mathbb{R}^d$. 

For answer prediction, we seek to minimize the cross entropy loss by $\mathcal{L}_{ans}=-\log softmax(W_zC+b_z)_l$, where $C$ is the combined representations of $\hat{O}$ and $H$. 

\paragraph{Logical Fact Regularization.}Since the subject, verb, and object in a fact should be closely related with some explicit relationships, we design a logical fact regularization technique to make the logical facts more of factual correctness.
Inspired by \cite{NIPS2013_1cecc7a7}, the embedding of the tail argument should be close to the embedding of the head argument plus a relation-related vector in the hidden representation space, i.e.,
    $v_{subject} + v_{predicate}\rightarrow v_{object}$. Specifically, given the hidden states of the sequence $h_i$ from the Transformer encoder. Regularization is defined as
\begin{equation}
     L_{lfr} = \sum_{k=1}^{m}(1-\cos(h_{sub_{k}}+h_{pred_{k}}, h_{obj_{k}})),
\end{equation}
   
where $m$ is the total number of logical fact triplets as well as the option and $k$ indicates the $k$-th fact triplet.

\subsection{Training Objective}
During training, the overall loss for answer prediction is $\mathcal{L} = \alpha\mathcal{L}_{ans} + \beta\mathcal{L}_{lfr}$,
where $\alpha$ and $\beta$ are two parameters. In our implementation, we set $\alpha=1.0$ and $\beta=0.5$.

\section{Experiments}
\label{others}

\subsection{Experimental Setup}
We conducted the experiments on three datasets. Two for specialized logical reasoning ability testing: ReClor \citep{yu2020reclor} and LogiQA \citep{ijcai2020-0501} and one for logical reasoning in dialogues: MuTual \citep{mutual}.

We take RoBERTa-large \citep{2019arXiv190711692L} and DeBERTa-xlarge \citep{he2020deberta} as our backbone models for convenient comparison with previous works. We also compare our model with previous baseline models as listed in Table \ref{related-work}. For more details on datasets and baseline models, one can refer to Appendix \ref{detail_datasets}.

The model is end-to-end trained and updated by Adam \citep{2014arxiv1412.6980k} optimizer with an overall learning rate of 8e-6 for ReClor and LogiQA, and 4e-6 for MuTual. The weight decay is $0.01$. We set the warm-up proportion during training to $0.1$. Graph encoders are implemented using DGL, an open-source lib of python. The layer number of the graph encoder is $2$ for ReClor and $3$ for LogiQA.  The maximum sequence length is 256 for LogiQA and MuTual, and 384 for ReClor. The model is trained for $10$ epochs with a total batch size of 16 and an overall dropout rate of $0.1$ on 4 NVIDIA Tesla V100 GPUs, which takes around 2 hours for ReClor and 4 hours for LogiQA.

\subsection{Results}    
Tables \ref{table:e2e} and \ref{mutual} show the results on ReClor, LogiQA, and MuTual, respectively. All the best results are shown in bold. We also provide variances of performance in Appendix \ref{variances} for reference. From the results, we have the following observations:

(i) Based on our implemented baseline model RoBERTa (basically consistent with public results), we observe dramatic improvements on both of the logical reasoning benchmarks, e.g., on ReClor test set, \textsc{Focal Reasoner} achieves an absolute improvement of $+4.2\%$ on dev set and $+3.3\%$ on the test set. \textsc{Focal Reasoner} also outperforms the prior best system LReasoner\footnote{The test results are from the official leaderboard \url{https://eval.ai/web/challenges/challenge-page/503/leaderboard/1347}.}, reaching $77.05\%$ on the EASY subset, and $44.64\%$ on the HARD subset. The superiority of the HARD subset indicates that our method is better at solving more complex questions that rely on reasoning over complex logical clues. The performance suggests that \textsc{Focal Reasoner} makes better use of logical structure inherent in the given context to perform reasoning than existing methods. Additionally, \textsc{Focal Reasoner} achieves consistent improvement on DeBERTa backbone, which indicates that our method is generally effective on stronger baselines.

(ii) Table \ref{class} specifies the accuracy of our model on the dev set of ReClor of different question types. Results show that our model can perform well on most of the question types, especially ``Strengthen'' and ``Weaken'', which generally involve statements such as ``which of the following weakens the conclusion?'' of negative semantics. This means that our model can well interpret the question type from the question statement and make the correct choice corresponding to the question, especially those with negation implications. Note that here our definition of ``negation statement'' is broader than traditional logic literature, which often contains words such as ``not'' and ``never''.

\begin{table}
\centering
\setlength{\belowcaptionskip}{5pt}
\vspace*{-3mm}
\small
\setlength{\tabcolsep}{16.0pt}
\begin{tabular}{lc}
    \toprule
    Model    & Accuracy \\
    \midrule
    \textsc{Focal Reasoner} & 66.8\scriptsize$\pm 0.13$ \\
    \midrule
    \textit{Supergraph Reasoning}  \\
    \quad w/o global edge  & 64.6\scriptsize$\pm 0.32$\\
    \quad w/o co-reference edges & 64.8\scriptsize$\pm 0.24$\\
    \quad w/o logical fact regularization & 64.2\scriptsize$\pm 0.12$ \\
    \quad w/o edge type & 63.7\scriptsize$\pm 0.19$ \\
    \midrule
    \textit{Interactions} \\
    \quad - interactions&  65.5\scriptsize$\pm 0.52$  \\
    \bottomrule
  \end{tabular}
  \caption{Ablation results on the ReClor dev set.}\label{ablation}
\vspace*{-5mm}
\end{table}

(iii) Our model outperforms the previous baseline models on ReClor without data augmentation (DA) and even performs on par with those with DA. On LogiQA dataset, \textsc{Focal Reasoner} obtains the best performance even taking DA into consideration. Given LogiQA is a more abstractive dataset than ReClor, we may infer that fact units could indeed better capture the logical relations inside the context, which leads to broader coverage of the knowledge. Combining with hierarchical modeling methods, we can further improve the performance. One can refer to Appendix \ref{interpretation} for a detailed example of the modeling process.

(iv) On the dialogue reasoning dataset MuTual, our model achieves substantial improvements compared with the RoBERTa-base LM.\footnote{Since there are no official results on RoBERTa-large LM, we use RoBERTa-base LM instead for consistency.} Focal Reasoner is able to generalize to a different domain of datasets beyond the logical reasoning inherent in texts. This verifies our model's generalizability on other downstream reasoning task settings.

(v) For the model complexity, our method basically keeps as simple as previous models like DAGN. Our model only has 414M parameters compared with 355M in the baseline RoBERTa, and 400M in DAGN which also employs GNN. This showcases effectiveness and simplicity.

\section{Analysis}\label{sec:analysis}

\begin{figure*}
\vspace{-3mm}
  \begin{center}
    \includegraphics[width=0.48\textwidth]{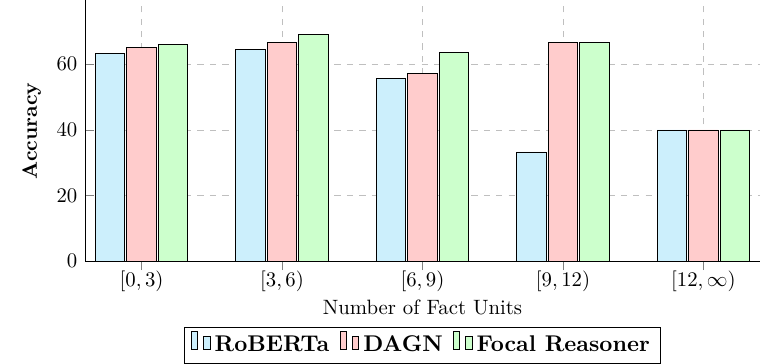}
    \hspace{3mm}
    \includegraphics[width=0.48\textwidth]{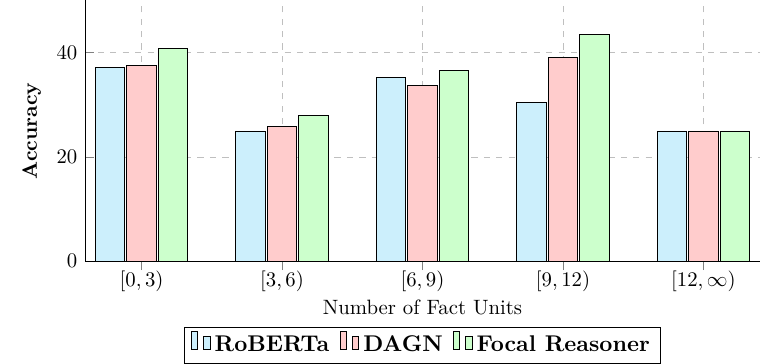}
  \end{center}
  \caption{Accuracy of models on a number of fact units on dev set of ReClor (left) and LogiQA (right).}
  \label{fig:analysis_1}
\end{figure*}

\subsection{Ablation Study}
To dive into the effectiveness of different components in \textsc{Focal Reasoner}, we conduct analysis by taking RoBERTa as the backbone of the ReClor dev set. Tables \ref{ablation} summarizes the results. 


\begin{table}
\vspace{-3mm}
\setlength{\belowcaptionskip}{2pt}
\small
\centering\centering\setlength{\tabcolsep}{5.6pt}
\begin{tabular}{lcccc}
\toprule
\multirow{2}{*}{Number} &
\multicolumn{2}{c}{ReClor} & \multicolumn{2}{c}{LogiQA} \\
\cmidrule{2-5}
& Train & Dev & Train & Dev  \\ 
\midrule
Fact Unit Argument&14,895& 1,665 &20,676  & 1,981 \\
Named Entity &9,495& 984  & 12,439 & 1,515 \\
\bottomrule
\end{tabular}
\caption{Statistics for fact unit entities and traditional named entities in datasets.}\label{tabel: num_fact_units}
\vspace{-3mm}
\end{table}

\paragraph{Supergraph Reasoning} The first key component is supergraph reasoning. We ablate the global atom which is initialized with the representation of concatenation of the question and each option. and erase all the edges connected with it. The results suggest that the global atom indeed betters message propagation, leveraging performance from $64.6\%$ to $66.8\%$. We also find that replacing the initial QA pair representation of the global atom with only question representation hurts the performance. In addition, without the logical fact regularization, the performance drops from $66.8\%$ to $64.2\%$, indicating its usefulness. For edge analysis, when (i) all edges are regarded as a single type rather than the originally designed 8 types in total and (ii) co-reference edges are removed, the accuracy drops to $63.7\%$ and $64.8\%$, respectively. It is proved that in our supergraph, edges link the fact units in reasonable manners, which properly uncovers the logical structures.

\paragraph{Interactions}

We further experimented with the query-option-interactions to see how it affects the performance. The results suggest that the features learned from the interaction process enhance the model. Considering that the logical relations between different options are a strong indicator of the right answer, this means that the model learns from a comparative reasoning strategy.

\subsection{Comparison with Alternative Fact Units}

\begin{table}
\centering
\setlength{\belowcaptionskip}{2pt}
\vspace*{-2mm}
\small
\setlength{\tabcolsep}{23.0pt}
\begin{tabular}{lc}
    \toprule
    Model    & Accuracy \\
    \midrule
    \textsc{Focal Reasoner} & 66.8\scriptsize$\pm 0.13$ \\
    \quad w/ named entity only & 62.8\scriptsize$\pm 0.26$ \\
    \quad w/ semantic role only & 62.2\scriptsize$\pm0.32$\\
    \bottomrule
  \end{tabular}
  \caption{Comparison about replacing fact units with commonsense knowledge such as named entities or semantic role labeling on the dev set of ReClor.}\label{comp_facts}
\vspace*{-5mm}
\end{table}

Apart from our syntactically constructed fact units, there are two other ways in different granularities for construction. We replace fact units with named entities that are used in previous works like \cite{Chen2019MultihopQA}. The statistics of fact units and named entities of ReClor and LogiQA are stated in Table \ref{tabel: num_fact_units}, from which we can infer that there are indeed more fact units than named entities. Thus using fact units can better incorporate the logical information within the context. When replacing all the fact units with named entities and leaving the model architecture unchanged, we can see from Table \ref{comp_facts} that it significantly decreases the performance. We also explore using semantic role labeling (SRL) in a similar way as in \cite{Zhong2020ReasoningOS}. SRL, leveraging much more complex information as well as computation complexity, fails to achieve performance as good as our original fact unit.


\subsection{Influence of Scale of Fact Units}

\begin{table}
\setlength{\belowcaptionskip}{5pt}
\vspace*{-2mm}
\centering\centering\setlength{\tabcolsep}{5.5pt}
\small
\begin{tabular}{lccccc}
\toprule
Dataset  & $[0,3)$ & $[3,6)$ & $[6,9)$ & $[9,12)$  & $[12, \infty)$ \\ 
\midrule
ReClor&37.2\%& 48.6\%  & 12.6\% & 0.6\%  & 1.2\% \\
LogiQA & 47.5\% & 37.5\%  & 10.9\% & 3.5\%  & 0.6\%\\
\bottomrule
\end{tabular}
\caption{Distribution of fact unit number on dev set of the training datasets.} 
\label{distribution}
\vspace*{-5mm}
\end{table}

To inspect the effects of the number of fact units, we split the original dev set of ReClor and LogiQA into $5$ subsets. The statistics of the fact unit distribution on the datasets are shown in Table \ref{distribution}. The numbers of fact units for most contexts in ReClor and LogiQA are in $[3,6)$ and $[0,3)$, respectively. Comparing the accuracies of RoBERTa-large baseline, prior SOTA LReasoner and our proposed \textsc{Focal Reasoner} in Figure \ref{fig:analysis_1}, our model outperforms baseline models on all the divided subsets, which demonstrates the effectiveness and robustness of our proposed method. Specifically, for ReClor, the performance of \textsc{Focal Reasoner} becomes more evident when the number of fact units locates in $[6,9)$, while for LogiQA, \textsc{Focal Reasoner} works better when the number of fact units locates in $[0,3)$ and $[9,12)$. The reason may lie in the difference in style of the two datasets. However, all the models including ours struggle when the number of fact units is above certain thresholds, i.e., the logical structure is more complicated, calling for better mechanisms to handle in the future.





\section{Conclusions}
In this work, we propose extracting a general form called ``fact unit'' to cover both commonsense and temporary knowledge units for logical reasoning. Our proposed \textsc{Focal Reasoner} not only better uncovers the logical structures within the context but also better captures the logical interactions between context and options. Experimental results verify the effectiveness of our method. 
\newpage

\paragraph{Limitations} There could be two limitations to the proposed method. First, as Figure 4 illustrates, when the documents become over-complicated, i.e., the number of facts units exceeds 12, our model would be less effective. Second, although we use parsers with high precision, propagated errors are inevitable. Designing systems without any external tools could be an interesting point in future studies.

\paragraph{Ethics Statement} Despite the recent success of pre-trained language models in machine reasoning, they still lack an explainability as to why and how they work in reasoning. \textsc{Focal Reasoner} takes the initial attempt to explore what kind of knowledge format is suitable and needed in logical reasoning of text. So far, we do not foresee any major risks or negative social impacts of our work. To encourage reproducibility, we provide source code in the supplementary material. The details of our method are described in Section 3. The hyperparameters for our model are discussed in Section 4.1. Variances of performances are shown in Appendix C. The ReClor and LogiQA datasets we experiment with are also publicly available at \url{https://github.com/yuweihao/reclor/releases/download/v1/reclor_data.zip} and \url{https://github.com/cylnlp/DialogSum} respectively. 

\bibliography{anthology,custom}

\begin{thebibliography}{63}
\expandafter\ifx\csname natexlab\endcsname\relax\def\natexlab#1{#1}\fi

\bibitem[{Back et~al.(2020)Back, Chinthakindi, Kedia, Lee, and
  Choo}]{back2020neurquri}
Seohyun Back, Sai~Chetan Chinthakindi, Akhil Kedia, Haejun Lee, and Jaegul
  Choo. 2020.
\newblock \href {https://openreview.net/forum?id=ryxgsCVYPr} {{NeurQuRI}:
  Neural question requirement inspector for answerability prediction in machine
  reading comprehension}.
\newblock In \emph{International Conference on Learning Representations}.

\bibitem[{Beck et~al.(2018{\natexlab{a}})Beck, Haffari, and
  Cohn}]{beck-etal-2018-graph}
Daniel Beck, Gholamreza Haffari, and Trevor Cohn. 2018{\natexlab{a}}.
\newblock \href {https://doi.org/10.18653/v1/P18-1026} {Graph-to-sequence
  learning using gated graph neural networks}.
\newblock In \emph{Proceedings of the 56th Annual Meeting of the Association
  for Computational Linguistics (Volume 1: Long Papers)}, pages 273--283,
  Melbourne, Australia. Association for Computational Linguistics.

\bibitem[{Beck et~al.(2018{\natexlab{b}})Beck, Haffari, and
  Cohn}]{beck2018graph}
Daniel Beck, Gholamreza Haffari, and Trevor Cohn. 2018{\natexlab{b}}.
\newblock Graph-to-sequence learning using gated graph neural networks.
\newblock \emph{arXiv preprint arXiv:1806.09835}.

\bibitem[{Bhagavatula et~al.(2019)Bhagavatula, Le~Bras, Malaviya, Sakaguchi,
  Holtzman, Rashkin, Downey, Yih, and Choi}]{bhagavatula2019abductive}
Chandra Bhagavatula, Ronan Le~Bras, Chaitanya Malaviya, Keisuke Sakaguchi, Ari
  Holtzman, Hannah Rashkin, Doug Downey, Wen-tau Yih, and Yejin Choi. 2019.
\newblock Abductive commonsense reasoning.
\newblock In \emph{International Conference on Learning Representations}.

\bibitem[{Bordes et~al.(2013)Bordes, Usunier, Garcia-Duran, Weston, and
  Yakhnenko}]{NIPS2013_1cecc7a7}
Antoine Bordes, Nicolas Usunier, Alberto Garcia-Duran, Jason Weston, and Oksana
  Yakhnenko. 2013.
\newblock \href
  {https://proceedings.neurips.cc/paper/2013/file/1cecc7a77928ca8133fa24680a88d2f9-Paper.pdf}
  {Translating embeddings for modeling multi-relational data}.
\newblock In \emph{Advances in Neural Information Processing Systems},
  volume~26. Curran Associates, Inc.

\bibitem[{Bowman et~al.(2015)Bowman, Angeli, Potts, and
  Manning}]{bowman2015large}
Samuel Bowman, Gabor Angeli, Christopher Potts, and Christopher~D Manning.
  2015.
\newblock A large annotated corpus for learning natural language inference.
\newblock In \emph{Proceedings of the 2015 Conference on Empirical Methods in
  Natural Language Processing}, pages 632--642.

\bibitem[{Chen et~al.(2016{\natexlab{a}})Chen, Bolton, and
  Manning}]{chen2016thorough}
Danqi Chen, Jason Bolton, and Christopher~D Manning. 2016{\natexlab{a}}.
\newblock A thorough examination of the cnn/daily mail reading comprehension
  task.
\newblock In \emph{Proceedings of the 54th Annual Meeting of the Association
  for Computational Linguistics (Volume 1: Long Papers)}, pages 2358--2367.

\bibitem[{Chen et~al.(2016{\natexlab{b}})Chen, Bolton, and
  Manning}]{chen-etal-2016-thorough}
Danqi Chen, Jason Bolton, and Christopher~D. Manning. 2016{\natexlab{b}}.
\newblock \href {https://doi.org/10.18653/v1/P16-1223} {A thorough examination
  of the {CNN}/{D}aily {M}ail reading comprehension task}.
\newblock In \emph{Proceedings of the 54th Annual Meeting of the Association
  for Computational Linguistics (Volume 1: Long Papers)}, pages 2358--2367,
  Berlin, Germany. Association for Computational Linguistics.

\bibitem[{Chen et~al.(2022)Chen, Zhang, and
  Zhao}]{chen-etal-2022-modeling-hierarchical}
Jialin Chen, Zhuosheng Zhang, and Hai Zhao. 2022.
\newblock \href {https://aclanthology.org/2022.coling-1.126} {Modeling
  hierarchical reasoning chains by linking discourse units and key phrases for
  reading comprehension}.
\newblock In \emph{Proceedings of the 29th International Conference on
  Computational Linguistics}, pages 1467--1479, Gyeongju, Republic of Korea.
  International Committee on Computational Linguistics.

\bibitem[{Chen et~al.(2019)Chen, Lin, and Durrett}]{Chen2019MultihopQA}
Jifan Chen, Shih-Ting Lin, and Greg Durrett. 2019.
\newblock Multi-hop question answering via reasoning chains.
\newblock \emph{ArXiv}, abs/1910.02610.

\bibitem[{Cui et~al.(2020)Cui, Wu, Liu, Zhang, and Zhou}]{mutual}
Leyang Cui, Yu~Wu, Shujie Liu, Yue Zhang, and Ming Zhou. 2020.
\newblock Mutual: A dataset for multi-turn dialogue reasoning.
\newblock In \emph{Proceedings of the 58th Conference of the Association for
  Computational Linguistics}. Association for Computational Linguistics.

\bibitem[{Cui et~al.(2017)Cui, Chen, Wei, Wang, Liu, and Hu}]{Cui2017Attention}
Yiming Cui, Zhipeng Chen, Si~Wei, Shijin Wang, Ting Liu, and Guoping Hu. 2017.
\newblock \href {https://www.aclweb.org/anthology/P17-1055.pdf}
  {Attention-over-attention neural networks for reading comprehension}.
\newblock In \emph{Proceedings of the 55th Annual Meeting of the Association
  for Computational Linguistics (Volume 1: Long Papers)}, pages 593--602.

\bibitem[{Davis and Marcus(2015)}]{davis2015commonsense}
Ernest Davis and Gary Marcus. 2015.
\newblock Commonsense reasoning and commonsense knowledge in artificial
  intelligence.
\newblock \emph{Communications of the ACM}, 58(9):92--103.

\bibitem[{Devlin et~al.(2019)Devlin, Chang, Lee, and
  Toutanova}]{devlin-etal-2019-bert}
Jacob Devlin, Ming-Wei Chang, Kenton Lee, and Kristina Toutanova. 2019.
\newblock \href {https://doi.org/10.18653/v1/N19-1423} {{BERT}: Pre-training of
  deep bidirectional transformers for language understanding}.
\newblock In \emph{Proceedings of the 2019 Conference of the North {A}merican
  Chapter of the Association for Computational Linguistics: Human Language
  Technologies, Volume 1 (Long and Short Papers)}, pages 4171--4186,
  Minneapolis, Minnesota. Association for Computational Linguistics.

\bibitem[{Dhingra et~al.(2017)Dhingra, Liu, Yang, Cohen, and
  Salakhutdinov}]{dhingra2017gated}
Bhuwan Dhingra, Hanxiao Liu, Zhilin Yang, William Cohen, and Ruslan
  Salakhutdinov. 2017.
\newblock \href {https://www.aclweb.org/anthology/P17-1168.pdf}
  {Gated-attention readers for text comprehension}.
\newblock In \emph{Proceedings of the 55th Annual Meeting of the Association
  for Computational Linguistics (Volume 1: Long Papers)}, pages 1832--1846.

\bibitem[{Dua et~al.(2019)Dua, Wang, Dasigi, Stanovsky, Singh, and
  Gardner}]{dua2019drop}
Dheeru Dua, Yizhong Wang, Pradeep Dasigi, Gabriel Stanovsky, Sameer Singh, and
  Matt Gardner. 2019.
\newblock Drop: A reading comprehension benchmark requiring discrete reasoning
  over paragraphs.
\newblock In \emph{Proceedings of the 2019 Conference of the North American
  Chapter of the Association for Computational Linguistics: Human Language
  Technologies, Volume 1 (Long and Short Papers)}, pages 2368--2378.

\bibitem[{Ettinger(2020)}]{ettinger2020bert}
Allyson Ettinger. 2020.
\newblock What bert is not: Lessons from a new suite of psycholinguistic
  diagnostics for language models.
\newblock \emph{Transactions of the Association for Computational Linguistics},
  8:34--48.

\bibitem[{Garcez et~al.(2015)Garcez, Besold, De~Raedt, F{\"o}ldiak, Hitzler,
  Icard, K{\"u}hnberger, Lamb, Miikkulainen, and Silver}]{garcez2015neural}
Artur~d'Avila Garcez, Tarek~R Besold, Luc De~Raedt, Peter F{\"o}ldiak, Pascal
  Hitzler, Thomas Icard, Kai-Uwe K{\"u}hnberger, Luis~C Lamb, Risto
  Miikkulainen, and Daniel~L Silver. 2015.
\newblock Neural-symbolic learning and reasoning: contributions and challenges.
\newblock In \emph{2015 AAAI Spring Symposium Series}.

\bibitem[{Garcez et~al.(2022)Garcez, Bader, Bowman, Lamb, de~Penning,
  Illuminoo, Poon, and Gerson~Zaverucha}]{garcez2022neural}
Artur~d’Avila Garcez, Sebastian Bader, Howard Bowman, Luis~C Lamb, Leo
  de~Penning, BV~Illuminoo, Hoifung Poon, and COPPE Gerson~Zaverucha. 2022.
\newblock Neural-symbolic learning and reasoning: A survey and interpretation.
\newblock \emph{Neuro-Symbolic Artificial Intelligence: The State of the Art},
  342:1.

\bibitem[{He et~al.(2020)He, Liu, Gao, and Chen}]{he2020deberta}
Pengcheng He, Xiaodong Liu, Jianfeng Gao, and Weizhu Chen. 2020.
\newblock Deberta: Decoding-enhanced bert with disentangled attention.
\newblock \emph{arXiv preprint arXiv:2006.03654}.

\bibitem[{Hermann et~al.(2015)Hermann, Ko{\v{c}}isk{\`y}, Grefenstette,
  Espeholt, Kay, Suleyman, and Blunsom}]{hermann2015teaching}
Karl~Moritz Hermann, Tom{\'a}{\v{s}} Ko{\v{c}}isk{\`y}, Edward Grefenstette,
  Lasse Espeholt, Will Kay, Mustafa Suleyman, and Phil Blunsom. 2015.
\newblock Teaching machines to read and comprehend.
\newblock In \emph{Proceedings of the 28th International Conference on Neural
  Information Processing Systems-Volume 1}, pages 1693--1701.

\bibitem[{Honnibal and Montani(2017)}]{spacy2}
Matthew Honnibal and Ines Montani. 2017.
\newblock {spaCy 2}: Natural language understanding with {B}loom embeddings,
  convolutional neural networks and incremental parsing.
\newblock To appear.

\bibitem[{Hu et~al.(2019)Hu, Wei, Peng, Huang, Yang, and Li}]{hu2019read}
Minghao Hu, Furu Wei, Yuxing Peng, Zhen Huang, Nan Yang, and Dongsheng Li.
  2019.
\newblock \href
  {https://wvvw.aaai.org/ojs/index.php/AAAI/article/view/4619/4497} {Read+
  verify: Machine reading comprehension with unanswerable questions}.
\newblock In \emph{Proceedings of the AAAI Conference on Artificial
  Intelligence}, volume~33, pages 6529--6537.

\bibitem[{Huang et~al.(2019)Huang, Le~Bras, Bhagavatula, and
  Choi}]{huang2019cosmos}
Lifu Huang, Ronan Le~Bras, Chandra Bhagavatula, and Yejin Choi. 2019.
\newblock Cosmos qa: Machine reading comprehension with contextual commonsense
  reasoning.
\newblock In \emph{Proceedings of the 2019 Conference on Empirical Methods in
  Natural Language Processing and the 9th International Joint Conference on
  Natural Language Processing (EMNLP-IJCNLP)}, pages 2391--2401.

\bibitem[{Huang et~al.(2021)Huang, Fang, Cao, Wang, and Liang}]{zhang2021video}
Yinya Huang, Meng Fang, Yu~Cao, Liwei Wang, and Xiaodan Liang. 2021.
\newblock {DAGN}: Discourse-aware graph network for logical reasoning.
\newblock In \emph{NAACL}.

\bibitem[{Jiao et~al.(2022)Jiao, Guo, Song, and Nie}]{jiao2022merit}
Fangkai Jiao, Yangyang Guo, Xuemeng Song, and Liqiang Nie. 2022.
\newblock Merit: Meta-path guided contrastive learning for logical reasoning.
\newblock \emph{arXiv preprint arXiv:2203.00357}.

\bibitem[{Kaushik and Lipton(2018)}]{kaushik-lipton-2018-much}
Divyansh Kaushik and Zachary~C. Lipton. 2018.
\newblock \href {https://doi.org/10.18653/v1/D18-1546} {How much reading does
  reading comprehension require? a critical investigation of popular
  benchmarks}.
\newblock In \emph{Proceedings of the 2018 Conference on Empirical Methods in
  Natural Language Processing}, pages 5010--5015, Brussels, Belgium.
  Association for Computational Linguistics.

\bibitem[{Kingma and Ba(2015)}]{2014arxiv1412.6980k}
Diederik~P. Kingma and Jimmy Ba. 2015.
\newblock \href {http://arxiv.org/abs/1412.6980} {Adam: {A} method for
  stochastic optimization}.
\newblock In \emph{3rd International Conference on Learning Representations,
  {ICLR} 2015, San Diego, CA, USA, May 7-9, 2015, Conference Track
  Proceedings}.

\bibitem[{Krishna et~al.(2020)Krishna, Summers, and Wies}]{krishna2020local}
Siddharth Krishna, Alexander~J Summers, and Thomas Wies. 2020.
\newblock Local reasoning for global graph properties.
\newblock In \emph{European Symposium on Programming}, pages 308--335.
  Springer, Cham.

\bibitem[{Lai et~al.(2017{\natexlab{a}})Lai, Xie, Liu, Yang, and
  Hovy}]{lai2017race}
Guokun Lai, Qizhe Xie, Hanxiao Liu, Yiming Yang, and Eduard Hovy.
  2017{\natexlab{a}}.
\newblock Race: Large-scale reading comprehension dataset from examinations.
\newblock In \emph{Proceedings of the 2017 Conference on Empirical Methods in
  Natural Language Processing}, pages 785--794.

\bibitem[{Lai et~al.(2017{\natexlab{b}})Lai, Xie, Liu, Yang, and
  Hovy}]{lai-etal-2017-race}
Guokun Lai, Qizhe Xie, Hanxiao Liu, Yiming Yang, and Eduard Hovy.
  2017{\natexlab{b}}.
\newblock \href {https://doi.org/10.18653/v1/D17-1082} {{RACE}: Large-scale
  {R}e{A}ding comprehension dataset from examinations}.
\newblock In \emph{Proceedings of the 2017 Conference on Empirical Methods in
  Natural Language Processing}, pages 785--794, Copenhagen, Denmark.
  Association for Computational Linguistics.

\bibitem[{Lai et~al.(2021)Lai, Zhang, Feng, Huang, and Zhao}]{lai2021machine}
Yuxuan Lai, Chen Zhang, Yansong Feng, Quzhe Huang, and Dongyan Zhao. 2021.
\newblock Why machine reading comprehension models learn shortcuts?
\newblock In \emph{Findings of the Association for Computational Linguistics:
  ACL-IJCNLP 2021}, pages 989--1002.

\bibitem[{Levi(1942)}]{levi-1942}
Friedrich~Wilhelm Levi. 1942.
\newblock \emph{Finite geometrical systems: six public lectues delivered in
  February, 1940, at the University of Calcutta}.
\newblock University of Calcutta.

\bibitem[{Liu et~al.(2020)Liu, Cui, Liu, Huang, Wang, and
  Zhang}]{ijcai2020-0501}
Jian Liu, Leyang Cui, Hanmeng Liu, Dandan Huang, Yile Wang, and Yue Zhang.
  2020.
\newblock \href {https://doi.org/10.24963/ijcai.2020/501} {Logiqa: A challenge
  dataset for machine reading comprehension with logical reasoning}.
\newblock In \emph{Proceedings of the Twenty-Ninth International Joint
  Conference on Artificial Intelligence, {IJCAI-20}}, pages 3622--3628.
  International Joint Conferences on Artificial Intelligence Organization.
\newblock Main track.

\bibitem[{{Liu} et~al.(2019){Liu}, {Ott}, {Goyal}, {Du}, {Joshi}, {Chen},
  {Levy}, {Lewis}, {Zettlemoyer}, and {Stoyanov}}]{2019arXiv190711692L}
Yinhan {Liu}, Myle {Ott}, Naman {Goyal}, Jingfei {Du}, Mandar {Joshi}, Danqi
  {Chen}, Omer {Levy}, Mike {Lewis}, Luke {Zettlemoyer}, and Veselin
  {Stoyanov}. 2019.
\newblock \href {http://arxiv.org/abs/1907.11692} {{RoBERTa: A Robustly
  Optimized BERT Pretraining Approach}}.
\newblock \emph{arXiv e-prints}, page arXiv:1907.11692.

\bibitem[{Lv et~al.(2020)Lv, Guo, Xu, Tang, Duan, Gong, Shou, Jiang, Cao, and
  Hu}]{lv2020graph}
Shangwen Lv, Daya Guo, Jingjing Xu, Duyu Tang, Nan Duan, Ming Gong, Linjun
  Shou, Daxin Jiang, Guihong Cao, and Songlin Hu. 2020.
\newblock Graph-based reasoning over heterogeneous external knowledge for
  commonsense question answering.
\newblock In \emph{Proceedings of the AAAI Conference on Artificial
  Intelligence}, volume~34, pages 8449--8456.

\bibitem[{Marcheggiani and Titov(2017)}]{marcheggiani-titov-2017-encoding}
Diego Marcheggiani and Ivan Titov. 2017.
\newblock \href {https://doi.org/10.18653/v1/D17-1159} {Encoding sentences with
  graph convolutional networks for semantic role labeling}.
\newblock In \emph{Proceedings of the 2017 Conference on Empirical Methods in
  Natural Language Processing}, pages 1506--1515, Copenhagen, Denmark.
  Association for Computational Linguistics.

\bibitem[{Nakashole and Mitchell(2014)}]{nakashole2014language}
Ndapandula Nakashole and Tom Mitchell. 2014.
\newblock Language-aware truth assessment of fact candidates.
\newblock In \emph{Proceedings of the 52nd Annual Meeting of the Association
  for Computational Linguistics (Volume 1: Long Papers)}, pages 1009--1019.

\bibitem[{Nie et~al.(2020)Nie, Williams, Dinan, Bansal, Weston, and
  Kiela}]{nie2020adversarial}
Yixin Nie, Adina Williams, Emily Dinan, Mohit Bansal, Jason Weston, and Douwe
  Kiela. 2020.
\newblock Adversarial nli: A new benchmark for natural language understanding.
\newblock In \emph{Proceedings of the 58th Annual Meeting of the Association
  for Computational Linguistics}, pages 4885--4901.

\bibitem[{Peterson(1997)}]{peterson1997fact}
Philip~L Peterson. 1997.
\newblock \emph{Fact proposition event}, volume~66.
\newblock Springer Science \& Business Media.

\bibitem[{Prasad et~al.(2008)Prasad, Dinesh, Lee, Miltsakaki, Robaldo, Joshi,
  and Webber}]{prasad2008penn}
Rashmi Prasad, Nikhil Dinesh, Alan Lee, Eleni Miltsakaki, Livio Robaldo,
  Aravind~K Joshi, and Bonnie~L Webber. 2008.
\newblock The penn discourse treebank 2.0.
\newblock In \emph{LREC}. Citeseer.

\bibitem[{Rajpurkar et~al.(2016)Rajpurkar, Zhang, Lopyrev, and
  Liang}]{rajpurkar-etal-2016-squad}
Pranav Rajpurkar, Jian Zhang, Konstantin Lopyrev, and Percy Liang. 2016.
\newblock \href {https://doi.org/10.18653/v1/D16-1264} {{SQ}u{AD}: 100,000+
  questions for machine comprehension of text}.
\newblock In \emph{Proceedings of the 2016 Conference on Empirical Methods in
  Natural Language Processing}, pages 2383--2392, Austin, Texas. Association
  for Computational Linguistics.

\bibitem[{Ran et~al.(2019)Ran, Li, Hu, and Zhou}]{ran2019option}
Qiu Ran, Peng Li, Weiwei Hu, and Jie Zhou. 2019.
\newblock Option comparison network for multiple-choice reading comprehension.
\newblock \emph{arXiv preprint arXiv:1903.03033}.

\bibitem[{Reichertz(2004)}]{reichertz20044}
Jo~Reichertz. 2004.
\newblock 4.3 abduction, deduction and induction in qualitative research.
\newblock \emph{A Companion to}, page 159.

\bibitem[{Ren and Leskovec(2020)}]{ren2020beta}
Hongyu Ren and Jure Leskovec. 2020.
\newblock Beta embeddings for multi-hop logical reasoning in knowledge graphs.
\newblock \emph{Advances in Neural Information Processing Systems}, 33.

\bibitem[{Rogers et~al.(2020)Rogers, Kovaleva, and
  Rumshisky}]{rogers2020primer}
Anna Rogers, Olga Kovaleva, and Anna Rumshisky. 2020.
\newblock A primer in bertology: What we know about how bert works.
\newblock \emph{Transactions of the Association for Computational Linguistics},
  8:842--866.

\bibitem[{Russell and Norvig(2002)}]{russell2002artificial}
Stuart Russell and Peter Norvig. 2002.
\newblock Artificial intelligence: a modern approach.

\bibitem[{Sachan and Xing(2016)}]{sachan2016machine}
Mrinmaya Sachan and Eric Xing. 2016.
\newblock Machine comprehension using rich semantic representations.
\newblock In \emph{Proceedings of the 54th Annual Meeting of the Association
  for Computational Linguistics (Volume 2: Short Papers)}, pages 486--492.

\bibitem[{Schlichtkrull et~al.(2018)Schlichtkrull, Kipf, Bloem, vanden Berg,
  Titov, and Welling}]{10.1007/978-3-319-93417-4_38}
Michael Schlichtkrull, Thomas~N. Kipf, Peter Bloem, Rianne vanden Berg, Ivan
  Titov, and Max Welling. 2018.
\newblock Modeling relational data with graph convolutional networks.
\newblock In \emph{The Semantic Web}, pages 593--607. Springer International
  Publishing.

\bibitem[{Seo et~al.(2017)Seo, Kembhavi, Farhadi, and
  Hajishirzi}]{Seo2016Bidirectional}
Minjoon Seo, Aniruddha Kembhavi, Ali Farhadi, and Hannaneh Hajishirzi. 2017.
\newblock \href {https://openreview.net/references/pdf?id=BkYnI4Wug}
  {Bidirectional attention flow for machine comprehension}.
\newblock In \emph{ICLR 2017}.

\bibitem[{Song et~al.(2018)Song, Wang, Yu, Zhang, Florian, and
  Gildea}]{song2018exploring}
Linfeng Song, Zhiguo Wang, Mo~Yu, Yue Zhang, Radu Florian, and Daniel Gildea.
  2018.
\newblock Exploring graph-structured passage representation for multi-hop
  reading comprehension with graph neural networks.
\newblock \emph{arXiv preprint arXiv:1809.02040}.

\bibitem[{Sugawara et~al.(2020)Sugawara, Stenetorp, Inui, and
  Aizawa}]{sugawara2020assessing}
Saku Sugawara, Pontus Stenetorp, Kentaro Inui, and Akiko Aizawa. 2020.
\newblock Assessing the benchmarking capacity of machine reading comprehension
  datasets.
\newblock In \emph{Proceedings of the AAAI Conference on Artificial
  Intelligence}, volume~34, pages 8918--8927.

\bibitem[{Talmor et~al.(2019)Talmor, Herzig, Lourie, and
  Berant}]{talmor2019commonsenseqa}
Alon Talmor, Jonathan Herzig, Nicholas Lourie, and Jonathan Berant. 2019.
\newblock Commonsenseqa: A question answering challenge targeting commonsense
  knowledge.
\newblock In \emph{NAACL-HLT (1)}.

\bibitem[{Wang et~al.(2021)Wang, Zhong, Tang, Wei, Fan, Jiang, Zhou, and
  Duan}]{wang2021logic}
Siyuan Wang, Wanjun Zhong, Duyu Tang, Zhongyu Wei, Zhihao Fan, Daxin Jiang,
  Ming Zhou, and Nan Duan. 2021.
\newblock Logic-driven context extension and data augmentation for logical
  reasoning of text.
\newblock \emph{arXiv preprint arXiv:2105.03659}.

\bibitem[{Williams et~al.(2018)Williams, Nangia, and
  Bowman}]{williams2018broad}
Adina Williams, Nikita Nangia, and Samuel Bowman. 2018.
\newblock A broad-coverage challenge corpus for sentence understanding through
  inference.
\newblock In \emph{Proceedings of the 2018 Conference of the North American
  Chapter of the Association for Computational Linguistics: Human Language
  Technologies, Volume 1 (Long Papers)}, pages 1112--1122.

\bibitem[{Yang et~al.(2018)Yang, Qi, Zhang, Bengio, Cohen, Salakhutdinov, and
  Manning}]{yang2018hotpotqa}
Zhilin Yang, Peng Qi, Saizheng Zhang, Yoshua Bengio, William Cohen, Ruslan
  Salakhutdinov, and Christopher~D Manning. 2018.
\newblock Hotpotqa: A dataset for diverse, explainable multi-hop question
  answering.
\newblock In \emph{Proceedings of the 2018 Conference on Empirical Methods in
  Natural Language Processing}, pages 2369--2380.

\bibitem[{Yasunaga et~al.(2021)Yasunaga, Ren, Bosselut, Liang, and
  Leskovec}]{yasunaga2021qagnn}
Michihiro Yasunaga, Hongyu Ren, Antoine Bosselut, Percy Liang, and Jure
  Leskovec. 2021.
\newblock Qa-gnn: Reasoning with language models and knowledge graphs for
  question answering.
\newblock In \emph{North American Chapter of the Association for Computational
  Linguistics (NAACL)}.

\bibitem[{Yu et~al.(2020)Yu, Jiang, Dong, and Feng}]{yu2020reclor}
Weihao Yu, Zihang Jiang, Yanfei Dong, and Jiashi Feng. 2020.
\newblock Reclor: A reading comprehension dataset requiring logical reasoning.
\newblock In \emph{International Conference on Learning Representations
  (ICLR)}.

\bibitem[{Zhang et~al.(2020{\natexlab{a}})Zhang, Wu, Zhou, Duan, Zhao, and
  Wang}]{zhang2019sg}
Zhuosheng Zhang, Yuwei Wu, Junru Zhou, Sufeng Duan, Hai Zhao, and Rui Wang.
  2020{\natexlab{a}}.
\newblock \href {https://arxiv.org/pdf/1908.05147.pdf} {{SG-Net}: Syntax-guided
  machine reading comprehension}.
\newblock In \emph{Proceedings of the Thirty-Fourth AAAI Conference on
  Artificial Intelligence (AAAI)}.

\bibitem[{Zhang et~al.(2020{\natexlab{b}})Zhang, Yang, and
  Zhao}]{zhang2020retrospective}
Zhuosheng Zhang, Junjie Yang, and Hai Zhao. 2020{\natexlab{b}}.
\newblock \href {https://arxiv.org/pdf/2001.09694.pdf} {Retrospective reader
  for machine reading comprehension}.
\newblock \emph{arXiv preprint arXiv:2001.09694}.

\bibitem[{{Zhong} et~al.(2021){Zhong}, {Wang}, {Tang}, {Xu}, {Guo}, {Wang},
  {Yin}, {Zhou}, and {Duan}}]{2021arXiv210406598Z}
Wanjun {Zhong}, Siyuan {Wang}, Duyu {Tang}, Zenan {Xu}, Daya {Guo}, Jiahai
  {Wang}, Jian {Yin}, Ming {Zhou}, and Nan {Duan}. 2021.
\newblock \href {http://arxiv.org/abs/2104.06598} {{AR-LSAT: Investigating
  Analytical Reasoning of Text}}.
\newblock \emph{arXiv e-prints}, page arXiv:2104.06598.

\bibitem[{Zhong et~al.(2020)Zhong, Xu, Tang, Xu, Duan, Zhou, Wang, and
  Yin}]{Zhong2020ReasoningOS}
Wanjun Zhong, Jingjing Xu, Duyu Tang, Zenan Xu, Nan Duan, M.~Zhou, Jiahai Wang,
  and Jian Yin. 2020.
\newblock Reasoning over semantic-level graph for fact checking.
\newblock In \emph{ACL}.

\bibitem[{Zhou et~al.(2020)Zhou, Duan, Liu, and Shum}]{ZHOU2020275}
Ming Zhou, Nan Duan, Shujie Liu, and Heung-Yeung Shum. 2020.
\newblock \href {https://doi.org/https://doi.org/10.1016/j.eng.2019.12.014}
  {Progress in neural nlp: Modeling, learning, and reasoning}.
\newblock \emph{Engineering}, 6(3):275--290.

\end{thebibliography}
\bibliographystyle{acl_natbib}

\appendix
\onecolumn
\section{Levi Graph Construction} \label{edge_type}

Levi graph transformation turns labeled edges into additional vertices. There are two types of edges in a traditional Levi graph: \textit{default} and \textit{reverse}. For example, an edge $(E1, R1, E2)$ in the original graph becomes $(E1, default, R1)$, $(R1, default, E2)$, $(R1, reverse, E1)$ and $(E2, reverse, R1)$.

However, the type of source and target vertices in the graph also matters \citep{beck2018graph}. Specifically, previous works use the same type of edge to pass information, which may reduce the effectiveness. Thus we propose to transform the \textit{default} edges into \textit{default-in} and \textit{default-out} edges, and the \textit{reverse} edges into \textit{reverse-in} and \textit{reverse-out} edges.

\section{Details for Datasets and Baseline Models}\label{detail_datasets}

In this section, we describe the datasets and baseline models used in the experiments.

\subsection{Datasets}
\paragraph{ReClor} ReClor contains 6,138 multiple-choice questions modified from standardized tests such as GMAT and LSAT, which are randomly split into train/dev/test sets with 4,638/500/1,000 samples respectively. It contains multiple logical reasoning types. The held-out test set is further divided into EASY and HARD subsets based on the performance of the BERT-based model \cite{devlin-etal-2019-bert}.
\paragraph{LogiQA} LogiQA consists of 8,678 multiple-choice questions collected from National Civil Servants Examinations of China and are manually translated into English by experts. The dataset is randomly split into train/dev/test sets with 7,376/651/651 samples correspondingly. LogiQA also contains various logical reasoning types.
\paragraph{MuTual}  MuTual has 8,860 dialogues annotated by linguist experts and high-quality annotators from Chinese high school English listening comprehension test data. It is randomly split into train/dev/test sets with 7,088/886/886 samples respectively. There more than $6$ types of reasoning abilities reflected in MuTual. MuTual$^{plus}$ is an advanced version, where one of the candidate responses is replaced by a safe response (e.g., \textit{``could you repeat that?''}) for each example.
\subsection{Baseline Models}
\paragraph{DAGN} explores passage-level discourse-aware clues used for solving logical reasoning QA. Specifically, they leverage discourse relations annotated in Penn Discourse TreeBank 2.0 (PDTB 2.0) \citep{prasad2008penn} and punctuation as the delimiters to split the context into elementary discourse units (EDUs). They are further organized into a logical graph and fed into a GNN to get the representation.

\paragraph{LReasoner} is a symbolic-driven framework for logical reasoning of text. It firstly identifies the logical symbols and expressions explicitly for the context and options based on manually designed rules. Then it performs logical inference over the expressions according to logical equivalence laws such as \textit{contraposition} \citep{russell2002artificial}. Finally, it verbalizes the expressions to match the answer.

\paragraph{HGN} proposes a holistic graph network (HGN) that deals with context at the discourse level and word level. Specifically, they leverage the EDUs and extract key phrases as the critical elements for the construction of HGN. Then they distinguish different edge types to link those nodes together.

\paragraph{MERIt} explores the self-supervised pre-training setting for logical reasoning. They employ the meta-path strategy to mine the potential logical structure in raw text in the form of relation triplets. Contrastive learning is also leveraged to further boost performance.

\paragraph{} Compared with previous baseline models, \textsc{Focal Reasoner} enjoys two major merits. (1) \textbf{Broader knowledge:} Compared with DAGN which uses sententious knowledge such as logical connectives (e.g., \textit{becaus}, \textit{however}), \textsc{Focal Reasoner} leverages a broader type of knowledge characterized by ``fact unit'', including commonsense knowledge and temporary knowledge. (2) \textbf{More transferable:} Compared with LReasoner which manually designs rules to extract logical patterns and perform logical reasoning in a symbolic way, Focal Reasoner is neural-based and manual-free, which is more generalizable to other datasets.







\section{Variances w.r.t Experiment Results} \label{variances}
In this section, we report the average and variances run on 5 random seeds for \textsc{Focal Reasoner} with different pre-trained language models.

\begin{table*}[htb]

\setlength{\belowcaptionskip}{5pt}
\caption{Experimental results for \textsc{Focal Reasoner} with average results and variances run no 5 random seeds.}
\centering\centering\setlength{\tabcolsep}{3.0pt}
\vskip 0.1in
\begin{tabular}{lcccccc}
\toprule
\multirow{2}{*}{Model} &
\multicolumn{4}{c}{ReClor} & \multicolumn{2}{c}{LogiQA}\\
\cmidrule{2-5}
\cmidrule{6-7}
 & Dev & Test & Test-E & Test-H & Dev & Test \\ 
\midrule
\textsc{Focal Reasoner}$_{\textup{RoBERTa}}$&\textbf{66.8}$\pm 0.13$ & \textbf{58.8}$\pm$0.14&\textbf{76.9}$\pm$0.16&\textbf{44.5}$\pm$0.12 & \textbf{41.0}$\pm$0.11&\textbf{40.3}$\pm$0.15\\
\textsc{Focal Reasoner}$_{\textup{DeBERTa}}$&\textbf{78.6}$\pm$0.18 & \textbf{73.2}$\pm$0.17&\textbf{86.2}$\pm$0.21&\textbf{62.9}$\pm$0.13 & \textbf{47.3}$\pm$0.16&\textbf{45.8}$\pm$0.17\\

\bottomrule
\end{tabular}

\end{table*}

\section{Case Study and Further Interpretation} \label{interpretation}
\begin{figure*}[h]
\centering
\vspace{-0.2cm}
\includegraphics[width=0.96\textwidth]{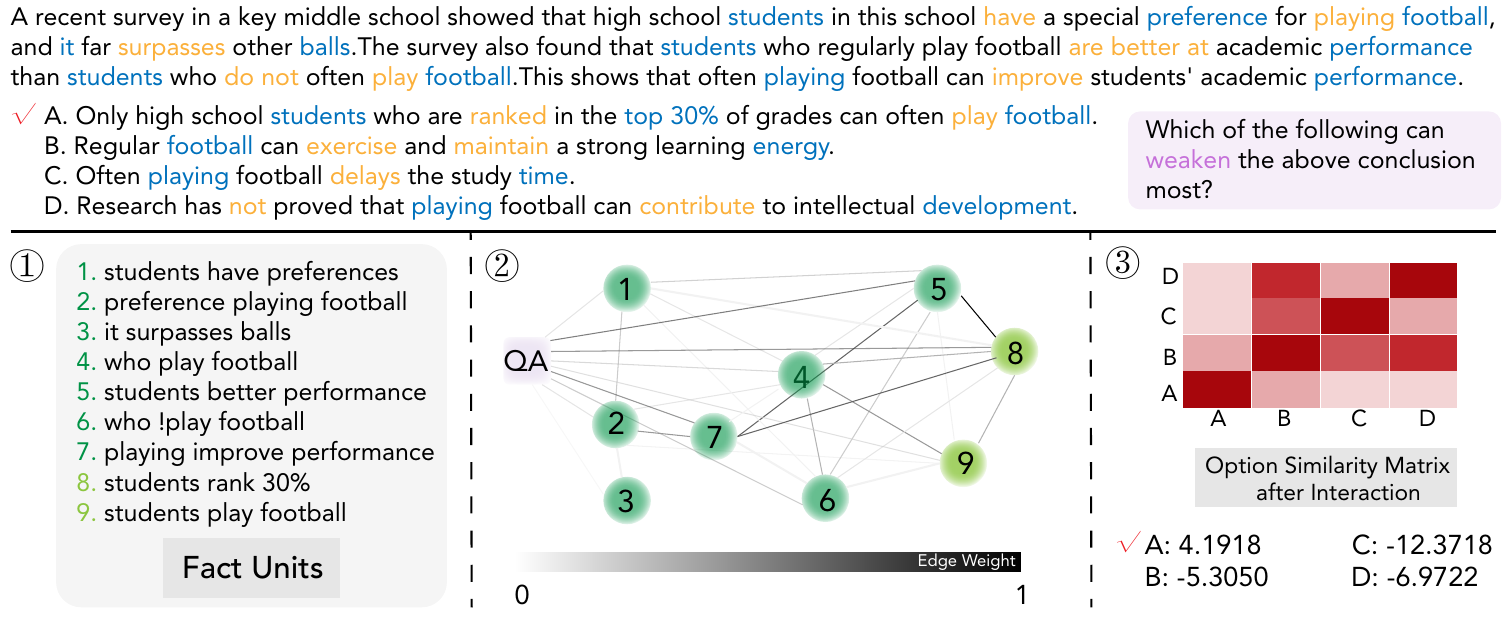}
\caption{An example of how our model reasons to get the final answer.}
\vspace{-0.2cm}
\label{case_study}
\end{figure*}

Figure \ref{case_study} illustrates our model's reasoning process by analyzing the node-to-node attention weights. We see that \textsc{Focal Reasoner} well bridges the reasoning process between context, question, and option. Specifically, in the graph, ``students rank 30\%'' attend strongly to ``playing improves performance''. Under the guidance of the question to select the option that weakens the statement and option interaction, our model is able to tell that ``students rank 30\% can play'' mostly undermines the conclusion that ``playing improves performance''.

\begin{figure*}[htb]
\centering
\includegraphics[width=0.96\textwidth]{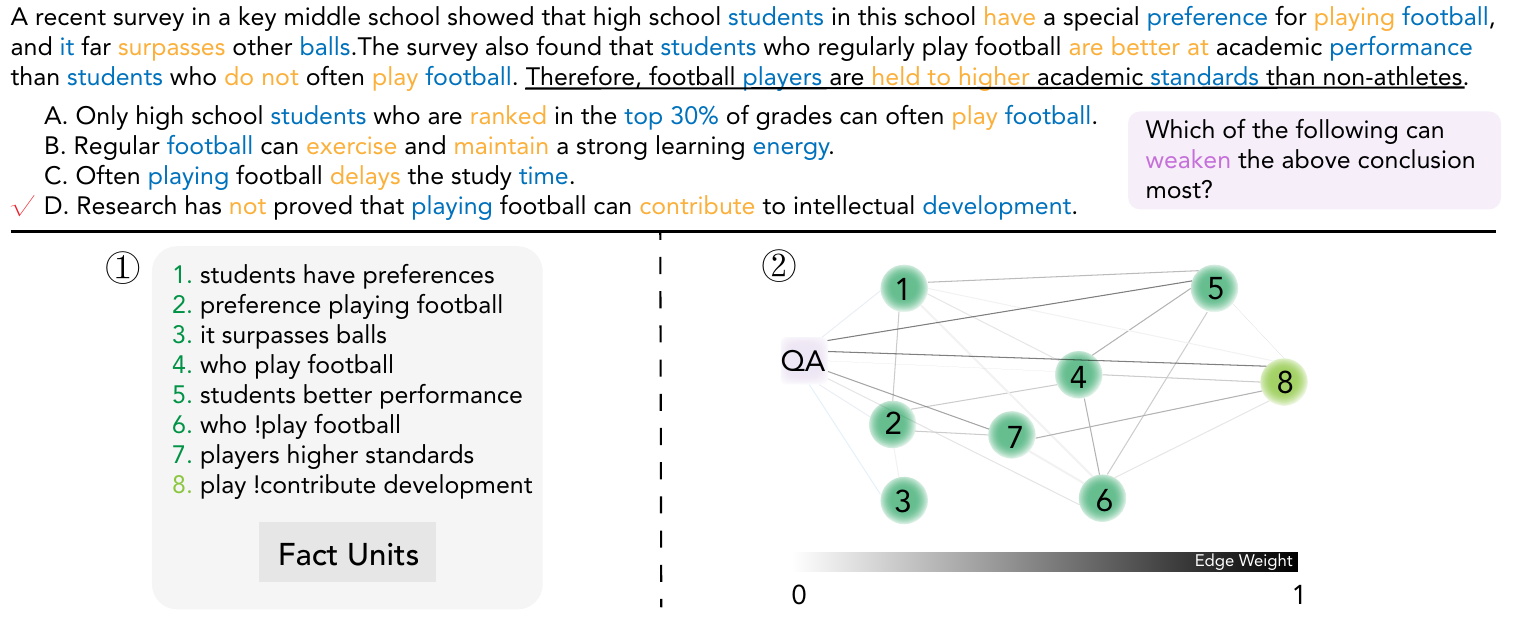}
\caption{An example of how our model reasons to get the final answer on the modified example.}
\label{case_study_m}
\end{figure*}

We change the example in Figure \ref{case_study} a bit. Specifically, we change the conclusion from ``playing football can improve students' academic performance'' to ``football players are held to higher academic standards than non-athletes'' as shown in Figure \ref{case_study_m}. We can observe that our model can select the relatively correct answer, which indicates that \textsc{Focal Reasoner} has some logical reasoning ability, instead of simple text matching and mining ability.

\end{document}